\documentclass{article}
\usepackage{ijcai18}

\usepackage{times}
\usepackage{xcolor}
\usepackage{soul}
\usepackage[utf8]{inputenc}
\usepackage[small]{caption}
\usepackage{times}
\usepackage{latexsym}
\usepackage{CJK}
\usepackage{url}
\usepackage{amsmath}
\usepackage{multirow}
\usepackage{booktabs}  
\usepackage{graphicx}
\usepackage{algorithm}  
\usepackage{algpseudocode} 
\usepackage{subfigure}
\usepackage{natbib}
\usepackage{pgfplots}
\usepackage{pgfplotstable}
\pgfplotsset{width=7.0cm,compat=1.5}
\usepackage{color,xcolor,framed}

\newcommand*{\affaddr}[1]{#1} 

\newcommand*{\email}[1]{\texttt{#1}}

\title{Decoding-History-Based Adaptive Control of Attention\\ for Neural Machine Translation}

\author{%
Junyang Lin, Shuming Ma, Qi Su, Xu Sun\\
\affaddr{MOE Key Laboratory of Computational Linguistics, Peking University}\\
\affaddr{School of Electronics Engineering and Computer Science, Peking University}\\
\affaddr{School of Foreign Languages, Peking University}\\
\email{\{linjunyang, shumingma, sukia, xusun\}@pku.edu.cn}
}

\begin{document}
\maketitle
\begin{CJK}{UTF8}{gbsn}
\begin{abstract}

Attention-based sequence-to-sequence model has proved successful in Neural Machine Translation (NMT). However, the attention without consideration of decoding history, which includes the past information in the decoder and the attention mechanism, often causes much repetition. To address this problem, we propose the decoding-history-based Adaptive Control of Attention (ACA) for the NMT model. ACA learns to control the attention by keeping track of the decoding history and the current information with a memory vector, so that the model can take the translated contents and the current information into consideration. Experiments on Chinese-English translation and the English-Vietnamese translation have demonstrated that our model significantly outperforms the strong baselines. The analysis shows that our model is capable of generating translation with less repetition and higher accuracy. The code will be available at \url{https://github.com/lancopku}
\end{abstract}

\section{Introduction}

With the development of Deep Learning, Neural Machine Translation (NMT) has demonstrated outstanding effects, and the sequence-to-sequence model (Seq2Seq) \citep{seq2seq} is the most commonly-used model in NMT. The attention mechanism \citep{attention,stanfordattention} is often used in the Seq2Seq model, and in many cases it can significantly improve the performance of the model. In translating, the decoder builds a language model on the target language for semantic coherence, the attention mechanism obtains the source-side information for the word generation at each time step.

However, the current source-side information that the attention mechanism acquires is often controversial to the translated contents because the attention has no knowledge of the translated contents. We present a typical example of the over-translation of the attention-based Seq2Seq model on the Chinese-English translation in Table \ref{example}. From the example, it can be found that the attention-based Seq2Seq generates the same phrase ``the Russian capital of Moscow'' for multiple times, causing much repetition.


With the motivation to tackle this problem, we propose our decoding-history-based Adaptive Control of Attention (ACA) for the attention-based Seq2Seq model. The mechanism controls the output of the attention based on the decoding history, including the past information in the decoder and the past alignment information in the attention mechanism. The computation of the attention requires the information of the memory vector, which is updated based on the decoding history, by manipulating the decoder output and the attention vector. With the help of the memory, the attention can be more adaptive to the translated contents so that the repetition in translation can be reduced.

Our main contributions include:
\begin{itemize}
\item We propose a decoding-history-based Adaptive Control of Attention for the NMT model, which tackles the conflict between the current attention and the decoding history so that the generation can be more adaptive to the translated contents;
\item Experiments on the Chinese-English translation and the English-Vietnamese translation show that our model outperforms the strong baselines, with the advantages of 3.61 BLEU score and 1.17 BLEU score over the best attention-based Seq2Seq model;
\item Compared with the strong baselines, the translation of our model is with less repetition and higher accuracy.
\end{itemize}


\begin{table}[t]
\small
\setlength{\tabcolsep}{3pt}
\centering
    \begin{tabular}{p{8cm}}
    \hline
    \textbf{Seq2Seq+Attention:} In\colorbox[rgb]{0.99,0.86,0.86}{the Russian capital of Moscow}, \colorbox[rgb]{0.99,0.86,0.86}{the} \colorbox[rgb]{0.99,0.86,0.86}{Russian capital of Moscow}of\colorbox[rgb]{0.99,0.86,0.86}{the Russian capital of Moscow} was killed this year because of the cold war, most of them were homeless and the elderly, including many people.\\
    \hline
    \textbf{Gold:} The temperatures in Moscow, capital of Russia, dropped to such low levels last night that even locals felt freezing cold. Six people died as a result, bringing up the death toll due to coldness this year to 239. Most of the dead were the homeless and the elderly, including many drunk.\\
    \hline  
    \end{tabular}
    \caption{An example of the translation of the conventional attention-based Seq2Seq model on the NIST 2003 Chinese-English translation task. The text highlighted indicates repetition.}
    \label{example}
\end{table}

\section{Attention-based Seq2Seq}
\begin{figure*}
     \centering
     \includegraphics[height=7.5cm]{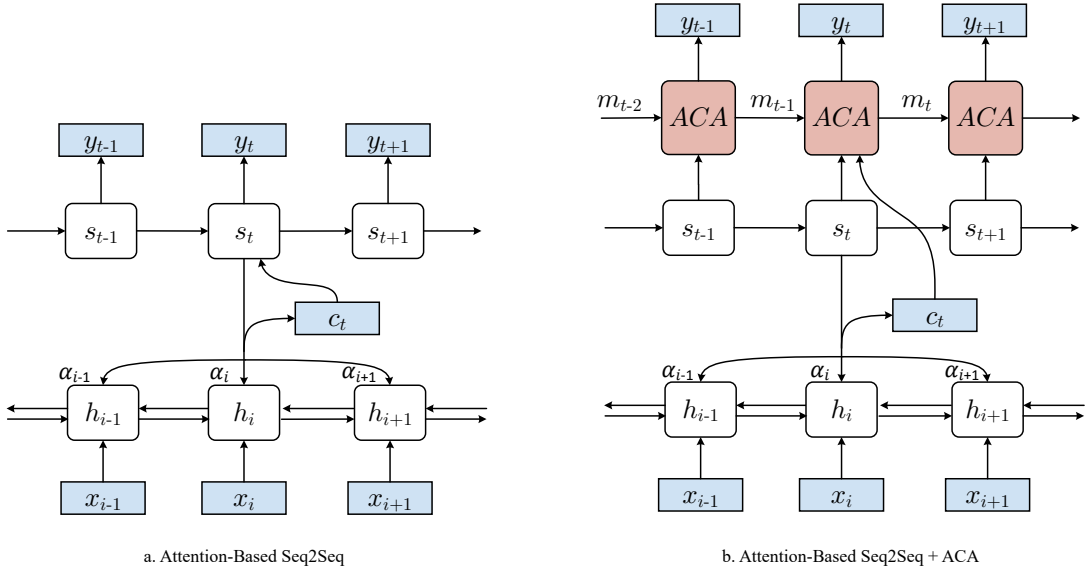}
     \caption{\textbf{Structure of the attention-based Seq2Seq and our Seq2Seq with ACA.} The left is the structure of the attention-based Seq2Seq model, and the right is the structure of our model, the attention-based Seq2Seq with the ACA.} 
     \label{fig1}
\end{figure*}

In Figure \ref{fig1}(a), we present a common type of attention-based Seq2Seq with RNN as its main component, and as we use LSTM in our model, we introduce the structure of LSTM in the following.

\subsection{Encoder}
As words are discrete units, the words in the source sequence should be sent through an embedding layer to become word embeddings as the input. On top of the embedding layer, the encoder turns the embeddings ${x = \{x_{1}, ..., x_{n}\}}$ into a sequence of encoder outputs ${h = \{h_{1}, ..., h_{n}\}}$ and sends out the final hidden state ${h_{n}}$ to the decoder.

The encoder in our model is a bidirectional LSTM, which is defined below:
\begin{align}
{f_{i}} &= {\sigma}({W_f[x_{i},h_{i-1}]} + {b_{f}}) \label{eq1}\\
{i_{i}} &= \sigma({W_i[x_{i},h_{i-1}]} + {b_{i}}) \label{eq2}\\
{o_{i}} &= {\sigma(W_o[x_{i},h_{i-1}] + b_{o})} \label{eq3}\\
{\tilde{C}_{i}} &= {tanh}({W_C[x_{i},h_{i-1}] + b_{C}}) \label{eq4}\\
{C_{i}} &= {f_{i}} \odot {C_{i-1} + i_{i}} \odot {\tilde{C}_{i}} \label{eq5}\\
{h_{i}} &= {o_{i}} \odot {tanh}({C_{i}}) \label{eq6}
\end{align}
where ${x_{i}}$ is the input word embedding at each time step from a minibatch of input sequences. LSTM consists of four gates, which collectively control the information flow from the last time step and the current time step. Bidirectional LSTM contains the same structure of LSTM, but it reads the input in two directions to generate two sequences of hidden states $\overrightarrow{{h}} \!=\! \{\overrightarrow{{h_{1}}}, \overrightarrow{{h_{2}}}, \overrightarrow{{h_{3}}}, ..., \overrightarrow{{h_{n}}}\}$ and $\overleftarrow{{h}} \!=\! \{\overleftarrow{{h_1}}, \overleftarrow{{h_2}}, \overleftarrow{{h_3}}, ... ,\overleftarrow{{h_n}}\}$, where:
\begin{align}
\overrightarrow{{h_{i}}} = {LSTM}({x_{i}}, \overrightarrow{{h_{i-1}}},{C_{i-1}}) \label{eq7}\\
\overleftarrow{{h_{i}}} = {LSTM}({x_{i}}, \overleftarrow{{h_{i-1}}}, {C_{i-1}}) \label{eq8}
\end{align}
The encoder outputs corresponding to each time step are concatenated as mentioned below:
\begin{align}
{h_i}\!=\![\overrightarrow{{h_i}};\overleftarrow{{h_i}}] \label{eq9} 
\end{align}


\subsection{Decoder}
The decoder is responsible for decoding the final state of the encoder ${h_{n}}$ to a new sequence ${y = \{y_{1}, ..., y_{m}\}}$. With the final encoder state as the initial state, the decoder is initialized to decode step by step, with a word embedding at each time step, until it generates the token representing the end-of-sentence mark.

For the decoder, we implement a unidirectional LSTM. The output of each time step is sent into a feed-forward neural network to be projected into the space of vocabulary ${Y} \in {R}^{\mid{Y}\mid \times {dim}}$. At each time step, the decoder generates a word ${y_{t}}$ by sampling from a distribution of the target vocabulary ${P_{vocab}}$, where: 
\begin{align}
{P_{vocab}} &= {softmax}({W_{o}v_{t}}) \label{predict}\\
{v_{t}} &= {g(W_{v}[c_{t};s_{t}])} \\
{s_{t}} &= {LSTM}({y_{t-1}, s_{t-1}}, {C_{t-1}}) \label{eq12}
\end{align}
where ${g(\cdot)}$ refers to non-linear activation function.

The global attention mechanism \citep{stanfordattention} is applied to the LSTM output ${s_{t}}$ and the encoders outputs ${h = \{h_{1},...,h_{n}\}}$ in order to obtain the global attention ${\alpha_{t,i}}$ and the context vector ${c_{t}}$, which is described in the following:
\begin{align}
{c_{t}} &= {\sum^{n}_{i=1} \alpha_{t,i}h_{i}} \label{eq13}\\
{\alpha_{t,i}} &= \frac{{exp}({e_{t,i}})}{{\sum_{j=1}^{n}}{exp}({e_{t,j}})} \label{eq14}\\
{e_{t,i}} &= {s_{t-1}^{\top}}{W_{a}h_{i}} \label{eq15}
\end{align}

\subsection{Training}

The training for the Seq2Seq model is usually based on maximum likelihood estimation. Given the parameters $\theta$ and source text $x$, the model generates a sequence $\tilde{y}$. The learning process is to minimize the negative log-likelihood between the generated text $\tilde{y}$ and reference $y$, which in our context is the sequence in target language for machine translation and summary for abstractive summarization:
\begin{align}
\mathcal{L} &= -\frac{1}{{N}}{\sum_{i=1}^{N}}{\sum_{t=1}^{T}}{p(y_{t}^{(i)}}|{\tilde{y}_{<t}^{(i)},x^{(i)}, \theta)} \label{eq22}
\end{align}
where the loss function is equivalent to maximizing the conditional probability of sequence $y$ given parameters $\theta$ and source sequence $x$.

\section{Adaptive Control of Attention}

As it is mentioned above, it is easy for the conventional attention-based Seq2Seq NMT models to suffer from generating incoherent texts due to the conflict between the attention mechanism and the decoding history. Based on the hypothesis, we propose our decoding-history-based Adaptive Control of Attention (ACA) mechanism to tackle the problem. Instead of sending the context vector ${c_{t}}$ directly to the output layer at each time step, we propose to update the attention with a recurrent memory that stores the information from the previous decoding time steps, so that the information from the attention mechanism can be controlled to be most beneficial to the whole generation. The memory updates itself at each time step with the information from the current decoder output and the current context vector, so that it can learn to remove unnecessary information and store important information at each time step. Moreover, it is responsible for restricting the information flow of the context vector in order to mitigate the conflict between attention and neural language model.

\begin{figure}
     \centering
     \includegraphics[height=3.0cm]{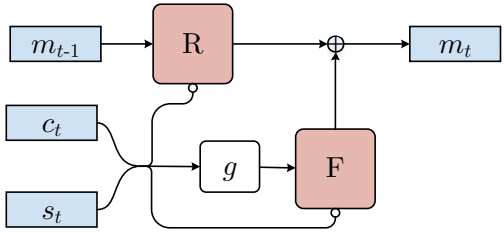}
     \caption{\textbf{Structure of the Recurrent Memory in the ACA.} ``R'' and ``F'' refer to the Remove and Feed operations, and ``g'' refers to the MLP.} 
     \label{fig2}
\end{figure}
\subsection{Recurrent Memory}

As our objective in this study is to build connection for the attention at the current time step with the decoding history, we implement the Recurrent Memory in the decoder for updating the context vector. The recurrent memory in the model is responsible for controlling the information flow of the attention mechanism, so that the effects of the attention mechanism can be connected with the previous decoding outputs from the RNN as well as the attention mechanism. Moreover, the memory should be updated at every decoding time step, so that it can reflect the development of the decoding history.

The memory ${m_{t}}$ is a representation vector at the decoding time step ${t}$, whose initialization ${m_{0}}$ is the last hidden state of the encoder, which is also the initial state for the RNN of the decoder. At each decoding time step ${t}$, the memory ${m_{t}}$ is updated with a \textbf{Remove-Feed} operation. The operation is based on the decision of the decoder output ${s_{t}}$ from the RNN and the context vector ${c_{t}}$ from the attention mechanism, so that the memory can observe the situation at the current time step and update itself with the guide of the current information. The structure of the Recurrent Memory is presented in Figure~\ref{fig2}.

At the beginning, the previous memory ${m_{t-1}}$ experiences a \textbf{Remove-Feed} operation. The decoder output ${s_{t}}$ and the context vector ${c_{t}}$ generate a Remove Gate ${r_{t}}$ to decide how to update the memory to be adaptive to the current decoding and a Feed Gate ${f_{t}}$ to decide how to update the memory with new information from the decoder and the attention:
\begin{align}
{r_{t}} &= {\sigma(g_{r}([s_{t};c_{t}]))}\\
{f_{t}} &= {\sigma(g_{f}([s_{t};c_{t}]))}
\end{align}
where ${g(\cdot)}$ refers to non-linear activation function to generate a vector of the hidden size.

Next, the previous memory ${m_{t-1}}$ passes through the gates and updates itself to be ${m_{t}}$ by removing information from the previous memory and adding new information from the decoder output and the context vector:
\begin{align}
{m_{t}} &= {(r_{t} \odot m_{t-1}) \oplus (f_{t} \odot g_{i}([s_{t};c_{t}]))} 
\end{align}

The Remove operation based on ${s_{t}}$ and ${c_{t}}$ can update the information stored in the memory based on the decoding and the attention at the current time step, so that the memory can be adaptive to the current decoding. The Feed operation based on the two same elements can provide the memory with the new information from the current time step so that the memory can store the repetition of the translated contents. Next, we introduce how the model makes use of the time-sensitive memory to improve the decoding.

\begin{figure}
     \centering
     \includegraphics[height=3.0cm]{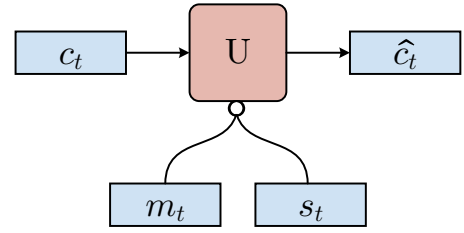}
     \caption{\textbf{Structure of Gated Control.} ``U'' refers to the Update Gate.} 
     \label{fig3}
\end{figure}
\subsection{Gated Control}

After the update, before entering the next time step, the memory ${m_{t}}$ collaborates with the decoder output ${s_{t}}$ to generate a gate for the context vector ${c_{t}}$. Therefore, the information from the attention mechanism is controlled by the information from the decoding history and the current state with the help of the updated memory. The detail operations are illustrated below.

Before entering the next time step, the current memory ${m_{t}}$ and the decoder output ${s_{t}}$ generate an Update Gate ${u_{t}}$ for the context vector ${c_{t}}$:
\begin{align}
{u_{t}} &= {\sigma(g_{u}([m_{t};s_{t}]))} \\
{\hat{c}_{t}} &= {u_{t} \odot c_{t}}
\end{align}
where ${\hat{c}_{t}}$ is the final context vector to be concatenated with the decoder output ${s_{t}}$ to generate the final output ${\hat{v}_{t}}$, which is given by:
\begin{align}
{\hat{v}_{t}} &= {g_{o}([s_{t};\hat{c}_{t}])}
\end{align}

In the final step of decoding, instead of sending ${v_{t}}$ to the output layer for the word prediction as mentioned in Equation \ref{predict}, the model sends ${\hat{v}_{t}}$, outputted from the ACA, for the prediction at each time step.

With the Gated Control, the context vector ${c_{t}}$ can be rectified based on the decoding history and the current information. The memory storing useful information of the partial translation can encourage to model to translate contents that are less repeated compared with the translated contents.Even if the source-side information in the context vector is in conflict with the decoding history, the conflict can be mitigated by the gate controlled by the memory.

\section{Experiment}

\begin{table*}[ht]
\centering
    \begin{tabular}{l|c|c|c|c|c|c|c}
    \hline
    Model  & MT-02 & MT-03 & MT-04 & MT-05 & MT-06 & MT-08 & Ave.\\ \hline\hline
    Moses \citep{lattice}& 33.19 & 32.43 & 34.14 & 31.47 & 30.81 & 23.85 & 31.04        \\ 
    RNNSearch \citep{lattice}& 34.68 & 33.08 & 35.32 & 31.42 & 31.61 & 23.58 & 31.76              \\ 
    Lattice \citep{lattice} & 35.94  &  34.32 &  36.50 &  32.40 & 32.77  &  24.84 &  32.95  \\
    Coverage \citep{Copy} & - & - & - & 32.73 & 32.47 & 25.23 & - \\
    Bi-Tree-LSTM \citep{bitreelstm} & 36.57 & 35.64 & 36.63 & 34.35 & 30.57 & - & -\\
    Mixed RNN \citep{mixedrnn} & 37.70 & 34.90 & 38.60 & 35.50 & 35.60 & - & - \\
    CPR \citep{PKI} & 33.84 & 31.18 & 33.26 & 30.67 & 29.63 & 22.38 & 29.72 \\
    POSTREG \citep{PKI} & 34.37 & 31.42 & 34.18 & 30.99 & 29.90 & 22.87 & 30.20 \\
    PKI \citep{PKI} & 36.10 & 33.64 & 36.48 & 33.08 & 32.90 & 24.63 & 32.51 \\
     \hline\hline
     Seq2Seq+Attention &  35.79 & 35.22  & 36.86 & 33.14  & 33.05 &  24.56 & 33.10 \\
    \textbf{+ACA} &  \textbf{40.25} & \textbf{38.31}  & \textbf{40.20} & \textbf{36.82}  &  \textbf{36.53} &  \textbf{28.14} & \textbf{36.71} \\
    \hline
    \end{tabular}
    \caption{Results of our model and the baselines (directly reported in the referred articles) on the Chinese-English translation, tested on the NIST Machine Translation tasks in 2003, 2004, 2005, 2006 with BLEU score. ``-'' means that the studies did not test the models on the corresponding datasets.}
    \label{cnen}
\end{table*}

\begin{table}[ht]
\centering
    \begin{tabular}{l|c}
    \hline
    Model & BLEU  \\ \hline\hline
    RNNSearch-1 \citep{luong2015stanford} &    23.30          \\ 
    RNNSearch-2 \citep{nplm} &     26.10                 \\ 
    LabelEmb \citep{labelemb} &     26.80                 \\
    NPMT \citep{nplm}&   27.69               \\ 
    NPMT+LM \citep{nplm}&   28.67               \\\hline\hline
     Seq2Seq+Attention &  26.93 \\
    \textbf{+ACA} &  \textbf{29.10} \\
    \hline
    \end{tabular}
    \caption{Results of our model and the baselines (directly reported in the referred articles) on the English-Vietnamese translation, tested on the TED tst2013 with the BLEU score.}
    \label{envi}
\end{table}

This section introduces the details of our experiments, including datasets, setups, baseline models as well as results.

\subsection{Datasets}
We evaluated our proposed model on the NIST translation task for Chinese-English translation and provided the analysis on the same task. Moreover, in order to evaluate the performance of our model on the low-resource translation, we also evaluated our model on the IWLST 2015 \citep{2015iwslt} for the English-Vietnamese translation task.

\textbf{Chinese-English Translation} For the NIST translation task, we trained our model on 1.25M sentence pairs extracted from LDC2002E18, LDC2003E07, LDC2003E14, Hansards portion of LDC2004T07, LDC2004T08 and LDC2005T06, with 27.9M Chinese words and 34.5M English words. Following \citet{lattice}, we validated our model on the dataset for the NIST 2005 translation task and tested our model on that for the NIST 2002, 2003, 2004, 2006, 2008 translation tasks. We used the most frequent 50,000 words for both the Chinese vocabulary and the English vocabulary. The evaluation metric is BLEU \citep{bleu}, and we calculated the case-insensitive NIST BLEU score with \texttt{multi-bleu.perl} provided by Moses\footnote{\url{http://www.statmt.org/moses/}.} .

\textbf{English-Vietnamese Translation} The data is from the translated TED talks, containing 133K training sentence pairs provided by the IWSLT 2015 Evaluation Campaign \citep{2015iwslt}. We followed the studies of \citet{nplm}, and used the same preprocessing as well as the validation set and the test set. The validation set is the TED tst2012 with 1553 sentences and the test set is the TED tst2013 with 1268 sentences. The English vocabulary is 17.7K words and the Vietnamese vocabulary is 7K words. The evaluation metric is also BLEU as mentioned above.

\subsection{Setting}
We implement the models using PyTorch, and the experiments are conducted on an NVIDIA 1080Ti GPU. Both the size of word embedding and hidden size are 512, and the batch size is 64. We use Adam optimizer \citep{KingmaBa2014} to train the model with the default setting $\beta_{1}=0.9$, $\beta_{2}=0.999$ and $\epsilon=1\times10^{-8}$, and we initialize the learning rate to $0.001$. 

Based on the performance on the development sets, we use a 3-layer LSTM as the encoder and a 2-layer LSTM as the decoder. Gradient clipping is applied so that the norm of the gradients cannot be larger than a constant, which is 10 in our experiments. Dropout is used with the dropout rate set to 0.2. 

Following \citet{multichannel}, we use beam search with a beam width of 10 to generate translation for the evaluation and test, and we normalize the log-likelihood scores by sentence length.

\subsection{Baselines}
In the following, we introduce our baseline models for the Chinese-English translation and the English-Vietnamese translation respectively.

\subsubsection{Chinese-English Translation}
Following \citet{lattice} and \citet{PKI}, we compare our model with the state-of-the-art NMT systems based on our implementation and the results directly reported in their articles, and we report the results of the baselines, Moses and RNNSearch from the study of \citet{lattice}.
\begin{itemize}
\item \textbf{Moses} An open source phrase-based translation system
with default configurations and a 4-gram language model trained on the training data for the target language;
\item \textbf{RNNSearch} An attention-based Seq2Seq with fine-tuned hyperparameters;
\item \textbf{Lattice} The Seq2Seq model with a Lattice-based RNN Encoder \citep{lattice};
\item \textbf{Bi-Tree-LSTM} A tree-coverage Seq2Seq model which lets the model depend on the source-side syntax \citep{bitreelstm};
\item \textbf{Mixed RNN} Extending RNNSearch with a mixed RNN as the encoder \citep{mixedrnn};
\item \textbf{CPR} Extending RNNSearch with a coverage penalty \citep{gnmt};
\item \textbf{POSTREG} Extending RNNSearch with posterior regularization with a constrained posterior set \citep{ganchev};
\item \textbf{PKI} Extending RNNSearch with posterior regularization to integrate prior knowledge \citep{PKI}.
\end{itemize}

\subsubsection{English-Vietnamese Translation}
Following \citet{luong2015stanford}, \citet{raffel} and \citet{nplm}, we compare our model with the state-of-the-art NMT models, and we present the results of the baseline directly reported in their studies.
\begin{itemize}
\item \textbf{RNNSearch-1} The attention-based Seq2Seq model by \citet{luong2015stanford};
\item \textbf{RNNSearch-2} The implementation of the attention-based Seq2Seq by \citet{nplm};
\item \textbf{LabelEmb} Extending RNNSearch with soft target representation \citep{labelemb};
\item \textbf{NPMT} The Neural Phrased-based Machine Translation model by \citet{nplm};
\item \textbf{NPMT-LM} On the basis of the NPMT, a trained 4th-order language model is added.
\end{itemize}


\begin{figure}[tb]
\begin{tikzpicture}
\begin{axis}[legend pos=north east, ybar, symbolic x coords={1-gram, 2-gram, 3-gram, 4-gram}, ylabel = {\% of the duplicates}, xtick=data]
\addlegendentry{w/o ACA}
\addplot 
coordinates{(1-gram, 18.7)(2-gram, 7.2)(3-gram, 4.3)(4-gram, 2.5)};

\addlegendentry{ACA}
\addplot 
coordinates{(1-gram, 16.6)(2-gram, 4.3)(3-gram, 2.2)(4-gram, 1.3)};

\end{axis}
\end{tikzpicture}
\caption{\textbf{Percentage of the duplicates at sentence level.} Tested on the NIST 2003 dataset. The red bar is the performance of our ACA, and the blue bar is the attention-based SeqSeq without ACA.}
\end{figure}
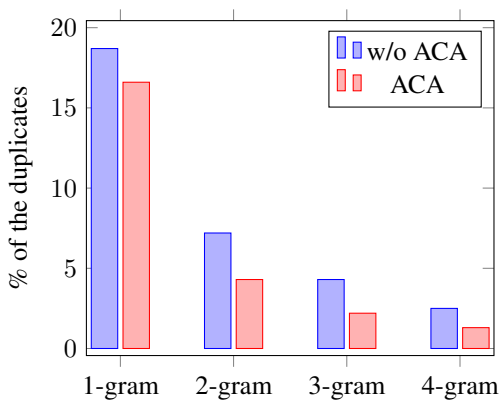

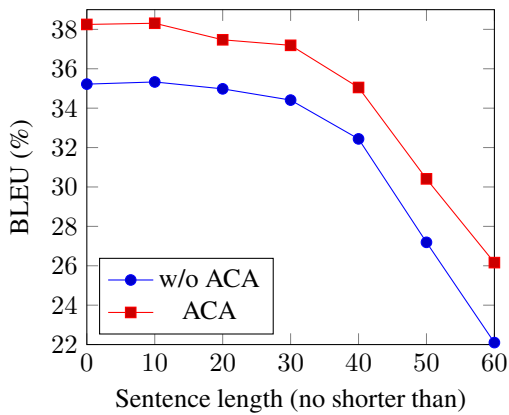
\begin{figure}[tb]
\begin{tikzpicture}
\begin{axis}[legend pos=south west, xlabel={Sentence length (no shorter than)}, ylabel = {BLEU (\%)}, xmin=0, xmax=60, ymin=22, ymax=39, xtick={0,10,20,30,40,50,60}, ytick={22,24,26,28,30,32,34,36,38,40}]
\addlegendentry{w/o ACA}
\addplot 
coordinates{(0, 35.22)(10, 35.33)(20, 34.98)(30, 34.41)(40, 32.44)(50, 27.19)(60, 22.10)};

\addlegendentry{ACA}
\addplot 
coordinates{
(0, 38.25)(10, 38.31)(20, 37.47)(30, 37.19)(40, 35.05)(50, 30.41)(60, 26.16)
};
\end{axis}
\end{tikzpicture}
\caption{\textbf{Performance on different sentence lengths.} Tested on the NIST 2003 dataset. The red line is the performance of our ACA, and the blue line is the attention-based SeqSeq without ACA.}
\end{figure}

\subsection{Results}
Table \ref{cnen} shows the overall results of the systems on the Chinese-English translation task. We compare our model with the strong baselines with their results directly reported in their articles. To facilitate fair comparison, we compare with the baselines that are trained on the same training set or slightly larger training set as reported in their articles. Many of the models are from the studies of the recent two years, which prove to be very strong baselines. The results have shown that for the six translation tasks, our ACA model has clear advantage over them, with 4.95 BLEU score over RNNSearch and 3.76 BLEU score over PKI, which proves that our model is effective.

Table \ref{envi} shows the overall results of the systems on the English Vietnamese translation. It can be found that on the low-resource translation, the ACA can also bring significant improvement for the attention-based Seq2Seq model, with the advantage of over 2.17 BLEU score over the strongest attention-based Seq2Seq and 1.41 BLEU score over the SOTA model NPMT. Moreover, compared with NPMT with a pretrained language model, our model is still better.

\subsection{Analysis}

In order to test whether our model can mitigate the problem of repetition in translation, we tested the repetition on the NIST 2003 dataset, following \citet{SeeEA2017}. We evaluated the proportion of the duplicates of 1-gram, 2-gram, 3-gram and 4-gram in each sentence and calculated the mean value. It can be found that at all levels, the translation of our model has less repetition. Moreover, the advantage of ours becomes clearer with the increase of the number of gram. Especially for the 4-gram, the proportion of duplicates of our model is almost only a half of that of the model without ACA. It is normal that there are repeating words in a sentence, but repeating 4-gram in most cases is unreasonable. Compared with the model without ACA, ACA can help the Seq2Seq model reduce unreasonable repetition and therefore mitigate the problem of over-translation by taking the decoding history into account.

Moreover, we choose the NIST 2003 Chinese-English translation dataset to test the performance of our model and the conventional attention-based Seq2Seq model without our ACA. We test the BLEU scores on sentences of length no shorter than 10, 20, 30, 40, 50, and 60. With the increase of length, the performance of both models decrease but our model ACA always has a clear advantage over the attention-based Seq2Seq. In our hypothesis, the model can adapt to the decoding history by improving the attention mechanism, so it is possible that it can perform better on the long-length sentence translation. Our analysis proves that the model can be more robust to translating sentences of diverse lengths.

\subsection{Translation Examples}
Table \ref{translation} shows two translation examples of our model on the NIST 2003 dataset, compared with the translation of the attention-based Seq2Seq model without ACA and the reference. It is obvious that both two translation examples of our example are similar to the references, outperforming those of the model without ACA, which has problems of repetition and meaning inconsistency. For the first sentence, the model without ACA generates repetition of ``cell phone users'' and misses the semantic unit ``top''. On the contrary, our translation is closer to literal translation, which is more faithful to the expression in the source. For the second example, it requires the model to reorder the translation since the name is followed by an adverbial phrase in the source. The complex and different structure in Chinese confused the model without ACA, which can only generate repetition of ``we are entering a new era''. With ACA, our model successfully reorders the translation by putting the name after the adverbial.

\begin{table}[t]
\small
\setlength{\tabcolsep}{3pt}
\centering
    \begin{tabular}{p{8cm}}
    \hline
    \textbf{Source:} 在此之前一年, 单单手机用户已跃居全球之冠 。 \\
    \hline
    \textbf{Reference:} The year before that, the number of mobile phone users alone already topped the world.\\
    \hline
    \textbf{Seq2Seq+Attention:} A year ago, cell phone users of cell phone users are already in the world.\\
    \hline
    \textbf{+ACA:} In the past year, cell phone users have leapt to the highest level in the world.\\
    \hline\\\hline
    \textbf{Source:} 佛莱文在谈及推行再生性能源策略已获致成功时表示: ``我们正进入一个新时代。''\\ 
    \hline
    \textbf{Reference:} Speaking about the success of promoting the strategy of renewable energies, Flavin said: ``we're entering a new era.''\\
    \hline
    \textbf{Seq2Seq+Attention:} ``We are entering a new era.'' ``We are ente-ring a new era.''\\
    \hline
    \textbf{+ACA:} Speaking on the success of the renewable energy strategy, Fortuyn said: ``we are entering a new era.''\\
    \hline
    \end{tabular}
    \caption{Two translation examples of our model, compared with the translation of the attention-based Seq2Seq model and the reference.}
    \label{translation}
\end{table}

\section{Related Work}


The studies of encoder-decoder framework \citep{Kalchbrenner,ChoEA2014,seq2seq} for this task launched the Neural Machine Translation (NMT). To improve the focus on the information in the encoder, \citet{attention} proposed the attention mechanism, which greatly improved the performance of the Seq2Seq model on NMT. Still, the attention mechanism suffers from prediction failure, and therefore, a number of studies were proposed to improve the mechanism, which also enhanced the performance of the NMT model \citep{stanfordattention,supervisedattention,mapattention,fengattention,Copy,micover,interactive,doublyattention,multichannel}. Some of them \citep{Copy,interactive} incorporated the previous attention into the current attention for better alignment, but none of them are based on the decoding history.

Besides improving attention mechanism for NMT, there are also some more effective neural networks. \citet{fairseq} turned the RNN-based model into CNN-based model, which greatly improves the computation speed. \citet{googleattention} removed the CNN and RNN and only used attention mechanism to build the model and showed outstanding performance. Also, some researches incorporated external knowledge in their systems and also achieved obvious improvement \citep{mixedrnn,bitreelstm}.


\section{Conclusion}
In conclusion, this paper proposes the decoding-history-based Adaptive Control of Attention (ACA) for the NMT model, which can transmit the significant information in the decoding history to control the output of the attention mechanism adaptively. Thus, the output of the attention mechanism is based on the the decoding history, including the past information in the RNN decoder as well as the alignment information in the attention mechanism. With this method, the conflict between the source-side information from the attention and the translated contents can be mitigated. Compared with the attention-based Seq2Seq model, our model captures more correct source information with the help of the decoding history and its translation behaves more adaptive to the past translation. Experiments on the Chinese-English translation and the English-Vietnamese translation all show that our model outperforms the strong baselines, which demonstrate the effectiveness of our model.

\end{CJK}
\nocite{MaEA2017,MaSun2017,SunEA2017,WeiEA2017,dnerre}
\bibliographystyle{named}
\bibliography{ijcai18}

\end{document}